# FPAN: Fine-grained and Progressive Attention Localization Network for Data Retrieval

Sijia Chen[1], Bin Song[1]*, Jie Guo[1], Xiaojiang Du[2], Mohsen Guizani[3]

[1]State Key Laboratory of Integrated Services Networks, Xidian University, Xi'an 710071, China
Email: sjchen@stu.xidian.edu.cn, bsong@mail.xidian.edu.cn, jguo@xidian.edu.cn

[2]Dept of Computer and Information Sciences, Temple University, Philadelphia, PA, USA,
Email: dxj@ieee.org

[3]Dept of Electrical and Computer Engineering, University of Idaho, Moscow, Idaho, USA
Email: mguizani@ieee.org

**Abstract** The Localization of the target object for data retrieval is a key issue in the Intelligent and Connected Transportation Systems (ICTS). However, due to lack of intelligence in the traditional transportation system, it can take tremendous resources to manually retrieve and locate the queried objects among a large number of images. In order to solve this issue, we propose an effective method to query-based object localization that uses artificial intelligence techniques to automatically locate the queried object in the complex background. The presented method is termed as Fine-grained and Progressive Attention Localization Network (FPAN), which uses an image and a queried object as input to accurately locate the target object in the image. Specifically, the fine-grained attention module is naturally embedded into each layer of the convolution neural network (CNN), thereby gradually suppressing the regions that are irrelevant to the queried object and eventually shrinking attention to the target area. We further employ top-down attentions fusion algorithm operated by a learnable cascade up-sampling structure to establish the connection between the attention map and the exact location of the queried object in the original image. Furthermore, the FPAN is trained by multi-task learning with box segmentation loss and cosine loss. At last, we conduct comprehensive experiments on both queried-based digit localization and object tracking with synthetic and benchmark datasets, respectively. The experimental results show that our algorithm is far superior to other algorithms in the synthesis datasets and outperforms most existing trackers on the OTB and VOT datasets.

## 1. Introduction

With the development of the transportation system connection network, the capacity of the data center is rapidly increasing, making efficient data retrieval an urgent need. However, due to the lack of intelligence, many of the retrieval tasks in the Intelligent and Connected Transportation Systems (ICTS) need to be done manually, resulting in inefficiencies and waste of resources. In all these data retrieval tasks in ICTS, the stable and effective queried object localization is the key task. As shown in Fig. 1, this well-known object localization by a query is of substantial importance for a wide range of applications in data retrieval such as intelligent monitoring, tracking, and vehicle positioning. However, the lack of intelligence and accuracy have restricted the development of data retireval system. In order to solve these problems, we focus on the intelligent localization technology, as shown in Fig. 1.

An important question about any object localization tasks is how to efficiently match the queried template in the image under arbitrary deformation of the target object [1]. The main challenge is to filter out the background and select some regions while learning the discrimination feature space and effective distance measurement so that we can distinguish the targets from candidate objects [2] .

The same problem also exists in the human visual search task. However, with the help of attention mechanism, human beings can always accurately locate the target object. That is, through more and more careful observation of local regions, humans gradually filter out irrelevant regions and tend to focus attention selectively on the target object in the image. This idea of gradually approaching the target area in a coarse-to-fine manner is explored in [3].

Attention mechanism has been largely studied in Neuroscience and Computational Neuroscience [4], [5], which also has a large impact on neural computation as we need to select the most pertinent piece of information, rather than using all available information at once [6]. Motivated by the properties of attention mechanism, many researchers have applied the similar attention mechanism to several computer vision tasks and have achieved great results, including image caption generation [7], object detection [6], object tracking [8], and object localization [9].

Inspired by the above results, several frameworks are proposed for attention-based visual search with neural networks. Seo et al [29] propose a framework that is trained to gradually suppress irrelevant regions in an input image via a progressive attentive process over multiple layers of a convolution neural network. However, these models can only use a fixed size of local regions to determine the weights of attention. At the same time, as the depth of the network increases, the location information is gradually lost, resulting in a large localization deviation at the top layer.

In this paper, we propose a unified framework to intelligently locate the queried object in the image, while also solve the above problems. In addition, the proposed net can be trained end-to-end directly on image-level supervisions.The model is an attention-based fully convolution neural network, referred to as Fine-grained and Progressive Attention Localization Network (FPAN), which attaches fine-grained attention to each layer of a progressive attention network to generate accuracy attention distribution and uses the learnable cascade up-sampling to refine the predicted spatial location of the queried object.

To the best of our knowledge, FPAN is the first method that automatically locating the queried object in the image with a unified network, as shown in Fig. 2. The main contributions of this work include three folds.

(1) The end-to-end unified deep progressive attention network for queried object localization, which can suppress irrelevant information layer by layer and learn to focus on the target object in the image.
(2) The fine-grained attention model to improve the accuracy and stability of the attention generation.
(3) The model uses the learnable cascade up-sampling to integrate the attention maps pyramid information to gradually restore the details of the target so that the queried target object can be accurately located in the original image, which also achieves image-level supervisions.

The rest of the paper is organized as follows. Section 2 reviews the related work. Section 3 introduces the proposed framework. Section 4 describes the training details of the framework. Section 5 provides the evaluation and analysis of the experiments, followed by the conclusion in Section 6.

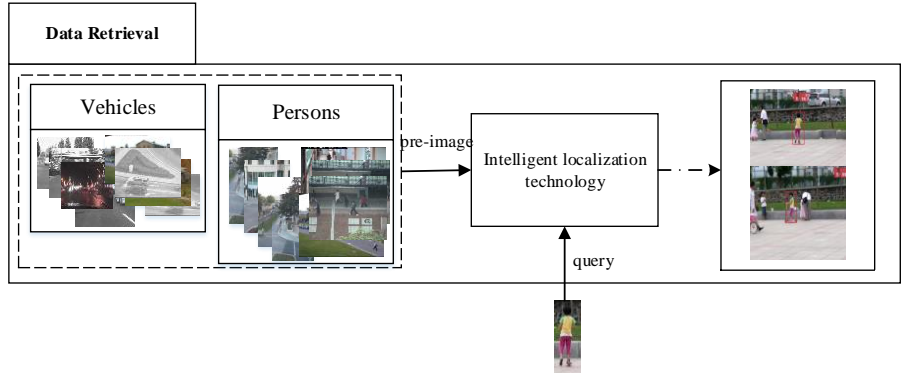

**Fig. 1.** The pipeline of queried object localization task in the data retrieval.

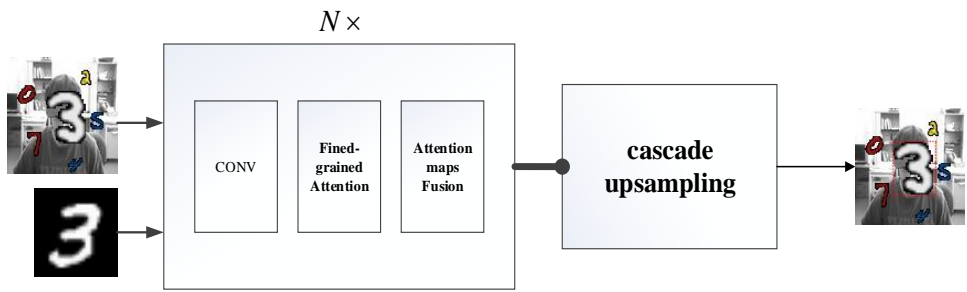

**Fig. 2.** The main structure of FPAN. FPAN accepts two inputs, one of which is an image need to be searched, and the other is a query. The structure mainly contains a N-layers progressive attention network used to attend to the target area and a learnable deconvolution operations used to locate the queried object in the original image. The target object contained in the image is finally accurate located with a tight package box.

2. **Related work**

Our work stems from the study of object localization, which mainly uses the template matching and localization algorithms. In addition, to improve the performance of the localization system, the visual attention model is introduced. To make this clear, in this section, the background information of object localization and visual attention is presented.

2.1. Object Localization

Locating a specific object in the image is a widely studied task. Among them, the template matching algorithm is the most used one. Meanwhile, more localization algorithms are proposed to improve the accuracy of the target location. Several papers (e.g., [66-71]) have studied related issues.

For template matching [1] [2], the common approach is a two stages process which localizes target object by similarity measurement after region proposal. The common approaches for region proposal are grouping methods [10] [11] and window scoring methods [12], [13]. After this, several methods [14] [15] are used to measure similarity. However, from the study in [3], [16], [17], [18], we can see that joint learning of region proposal and detection can achieve better results. Especially, Amir Ghodrati et al [3] propose a method for proposing object location with CNN-based features and also refine the boxes with an inverse cascade operation. One of the most important conclusions in their research is that different activation layers of CNN can give complementary information for object localization. J. Dai et al [18] put forward to a region-based detector that uses the fully convolutional network with all computation shared on the entire image to improve efficiency. The research in [18] uses CNN-based representation to capture discriminative appearance factors and exhibits localization sensitivity which is essential for accuracy object localization.

The research of object localization mainly focuses on the regression model [16] or the specially designed network [19], [20]. Additional, integrating the discriminative feature extraction and localization into a unified network for joint learning is often a better solution [21], [22], [23]. Among them, Zhou et al [22] shed light on the properties of global average pooling layer that enables the CNN to have remarkable localization ability despite being trained on image level labels. They use global average pooling to retain the spatial information of the network and establish the association between feature maps and categories as a class activation map so that the object can be located more accurately. Fu et al [23] propose deconvolution single shot detector (DSSD) as so to introduce more advanced contextual information and improve performance.

2.2. Visual Attention

Behavioral experiments have shown that the human visual system makes extensive use of contextual information for facilitating object search in natural scenes. Motivated by human perception, modeling eye-movements [24] during search developed for understanding how humans select relevant information and structure behavior in real time. The TVA-based model [25] controls attention system by low-level static and dynamic visual features of the environment (bottom-up), medium-level visual features of proto-objects and the task (top-down). The RBM-based attention model [26] make the model can be learned by the location, orientation, scale and speed of the attended object.

With the development of Convolutional Neural Network (CNNs), the efficient approaches that combine attention model with this technique have yielded great results in visual detection and localization tasks, such as [6], [8], [27], [28]. The hard attention proposed in [6] recognizes objects through a series of glimpses, which can also attend to the target region. The soft attention [27] learns which parts in the frames are relevant and recognizes the human actions. With a soft attention framework, the RATM [8] performs well on several object tracking datasets including moving digits and the KTH dataset. Specifically, the attention net in [28] provides quantized weak directions pointing a target object and the ensemble of iterative predictions converges to an accurate object boundary box. The PANNet [29] is a novel attention model which can accurately attend to target objects of various scales and shapes in images. The net learns to capture semantic association between images and query while constructs function that maps information to attention value.

3. Fine-grained attention and progressive attention model

Given an image $x$ and a queried object $q$, we aim to determine the exact location of $q$ in $x$. Let $z = \{x, q\}$ represent an image and query pair. Then the ideal feature extraction function which extracts the most distinguishing features $\{f, f_q\}$ from the pair can be denoted as $F$. Meanwhile, a desirable retrieval function should be able to separate the target object from the background in $x$ and get the location of the queried object. Here, we consider the extraction of features and the retrieval of an object as two processes that interact with each other. To jointly optimize the two processes, we propose an iterative localization scheme, which has an explicit form, i.e.

$$\begin{cases} f_q^{l+1} = F^{l+1}(f_q^l) \\ f^{l+1} = F^{l+1}\left(Att^l \odot f^l\right) \\ Att_k^l = T_k^l(f^l, f_q^l) \\ Att^l = \Omega^l(Att_1^l, \dots, Att_P^l) \\ loc = \Theta(Att^1, Att^2, \dots, Att^L) \end{cases} \quad (1)$$

As shown in the above formula, the proposed method is an iterative algorithm with a coarse-to-fine manner, which is induced by fine-grained attention generation function T, attention maps fusion mechanism $\Omega$, cascade up-sampling operation $\Theta$. That is, in each layer of the convolution neural network, the multi-scale local image features and the main features of the query patch are extracted so that T can be used to generate fine-grained attention maps. Then, $\Omega$ integrates these attentions map so as to output the optimal attention distribution at the current layer. Furthermore, the image features map is multiplied to the attention map channel-wise so as to forward to next layer. Finally, $\Theta$ is used to determine the exact location of the queried object in the original image, which also facilitates image-level supervisions. It should be note that F is the pre-trained deep convolution neural network used in most of the computer vision tasks [16], [22].

The proposed algorithm utilizes the idea of progressive attention to determine the location of the queried object in the image and naturally combines it with the convolution neural network (CNN), which is illustrated in Fig. 3.

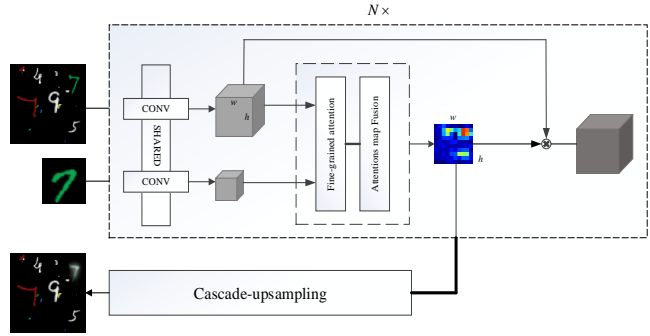

**Fig. 3.** The structure of FPAN.

3.1. Allocation of attention

Motivated by Paul et.al [29] who uses the text query as input, we use the features of the image and the queried object as input, and propose the following formulation:

$$\begin{cases} \bar{f}_q^l = E_q^l(f_q^l) \\ \varphi^l = G^l(f^l, \bar{f}_q^l) \\ s^l = T^l(\varphi^l; \theta_k^l) \end{cases} \quad (2)$$

where $f^l \in R^{W^l \times H^l \times C^l}$ and $f_q^l \in R^{W^l \times H^l \times C^l}$ are the feature maps generated from a block of convolution operations, $\bar{f}_q^l \in R^{C^l}$ is the main feature of the queried object encoded from $f_q^l$. It is worth noting that the image and the queried object share the same feature extraction operation as shown in Fig. 3.

As shown in Eq. (2), there are three functions that play a major role in the allocation of attention map. That is, $E_q^l$ encodes the main features of the object, $G^l$ establishes the association between the encoded feature and the image features, and $T^l$ assigns attention to the various regions.

Given the queried object with the features map $f_q^l \in R^{W_q^l \times H_q^l \times C^l}$ where each channel is donated as $(f_q^l)_{C_i^l}^{W_q^l \times H_q^l}$, our task is to extract the main features instead of generating high-level abstraction features. Inspired by the depth-wise separable convolution proposed in [30], which divides the convolution into two steps, named depth-wise convolution and pointwise convolution, respectively. Here, we remove the pointwise convolution and propose a new depth-wise convolution operation that extracts the main features specifically for subsequently feature similarity matching.

In addition, each channel represents a possible feature [31]. To this end, we define a convolution operation that treats each channel individually which can be donated as *CSCONV*, as shown in Eq. (3):

$$CSCONV(f) : \{f_k' : f_k' = f_k * \text{kernel}_k; f_k \in f, k \in K\} \quad (3)$$

As shown above, each channel of the feature maps has separate convolution kernels. The contribution of this operation is to encode the features expressed in each channel without changing its meaning, while greatly reduces the number of parameters.

We define the encoding function $E_q^l$ to be the cascaded *CSCONV* operations that act on the input feature map $f_q^l$. More formally:

$$\begin{cases} \bar{f}_q^l = E_q^l(f_q^l) = CSCONV\left(CSCONV\left(\ldots CSCONV(f_q^l)\right)\right) \\ \bar{f}_q^l \in R^{1 \times 1 \times C^l} \end{cases} \quad (4)$$

For each channel $(f_q^l)_{C_i^l}^{W_q^l \times R_q^l}$, the cascade *CSCONV* operations eventually output the corresponding $(f_q^l)_{C_i^l}^{1 \times 1}$, which is the mapping of the features map to the embeddable encoded feature known as $R^{W_q^l \times R_q^l} \to R^{1 \times 1}$.

By using the function $E_q^l$, we get the embeddable features vector $\bar{f}_q^l$ of the queried object which can be used to generate attention map.

$G^l$ aims to build a fusion feature map $\varphi^l$ for the input $\bar{f}^l$ and $\bar{f}_q^l$, where $\bar{f}^l$ is generated by $f^l$ with additional processing. Specifically, our goal is to design the $\bar{f}^l$ such that:
(1) $\bar{f}^l(z)$ for position z in each channel carries information from a local region of $f^l$ centered at location z.
(2) $\bar{f}^l$ should have the similar meaning as $\bar{f}_q^l$ in channel-wise..

These two properties can be obtained by defining an operation that is formulated as:

$$f^l \xrightarrow{CSCONV} \bar{f}^l \in R^{W^l \times H^l \times C^l} \quad (5)$$

Concretely, the stride of convolution is set to one so that $\bar{f}^l$ has same size as $f^l$.

Given $\bar{f}^l$ and $\bar{f}_q^l$, the $G^l$ for fusing them at corresponding location z at each channel $C_i^l$ is defined as:

$$[\varphi^l]_{C_i^l}^{W^l \times H^l}(z) = [\bar{f}^l]_{C_i^l}^{W^l \times H^l}(z) + [\bar{f}_q^l]_{C_i^l}^{W^l \times H^l} \quad (6)$$

As for the generation of attention map, we use the same operation defined in PAN [29]. In short, the operation $T^l$ that mapping the space of integrated features to attention values can be donated as $[\varphi^l]^{W^l \times H^l \times C^l} \to s^l \in R^{W^l \times H^l}$.

3.2. Fine-grained attentions model

In order to increase the accuracy and robustness of attention generation, we propose fine-grained attention model, which mainly includes two parts. On the one hand, multi-scale local contextual features are used to get $\bar{f}^l$. On the other hand, the final attention map is actually the result of ensemble various attentions maps. This model emphasizes on multi-size local information and ensemble of attention maps, which naturally leads to fine-grained attention architecture shown in Fig. 4.

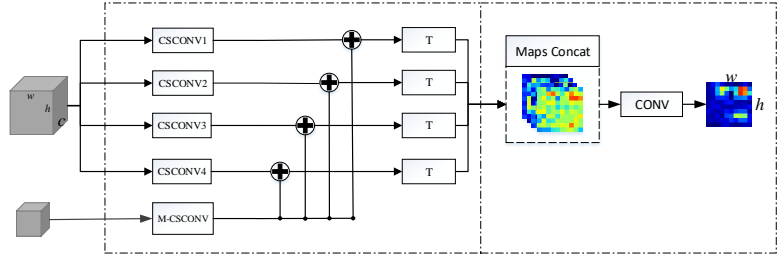

**Fig. 4.** Architecture of the generation of fine-grained attention.

3.2.1. Multi-scale local context information

As shown in Section 3.1, $\bar{f}^l$ is indispensable for the whole system because each of its position $[\bar{f}^l]_{i,j}$ is composed of corresponding spatially adjacent features in $f^l$.

However, using a single size of local information can cause several problems. 1) Difficulty in size selection. 2) Insufficient use of local information. 3) Only a single attention map can be generated, resulting in a low robustness of the system to abnormalities.

As the study in [32], increasing the width of the network can significantly improve the performance of the network. Therefore, in the structure $f^l \xrightarrow{CSCONV} \bar{f}^l$, we increase the width of the network, that is, multi-size $CSCONV$ is used to overcome these problems. The final structure is as follows:

$$\begin{cases} CSCONV_n(f): \{f'_k: f'_k = f_k * \text{kernel}^n_k; f_k \in f, k \in K\} \\ f^l \xrightarrow{CVCONV_n} (\bar{f}^l)^{CSCONV_n} \\ \overline{F^l_f} = \{(\bar{f}^l)^{CSCONV_1}, (\bar{f}^l)^{CSCONV_2}, \dots, (\bar{f}^l)^{CSCONV_P}\} \end{cases} \quad (7)$$

The framework produces a sequence of $\bar{f}^l$, represented by $\overline{F^l_f}$, each element $(\bar{f}^l)^{CSCONV_n}$ is generated by the $CSCONV$ with specific convolution type. Then, with $\overline{F^l_f}$ and $\bar{f}^l_q$ as inputs, we can get $\Psi^l = \{(\varphi^l)^1, (\varphi^l)^2, \dots, (\varphi^l)^P\}$ with each $(\varphi^l)^n$ generated by $G^l\left((\bar{f}^l)^{CSCONV_n}, \bar{f}^l_q\right)$.

3.2.2. Ensemble of attention maps

When there are multiple integrated feature maps available, that is $\Psi^l$, we can use $T^l$ to generate a series of attention maps which can be donated as $S^l: \{(s^l)^k: T^l_k((\varphi^l)^k); (\varphi^l)^k \in \Psi^l\}$. Then, given varieties of attention maps available, obtained by function $T^l$ with different parameters and different input, it is possible to use ensemble learning [33] to integrate these results.

Stacking [34] is an important way in ensemble learning that involves training a learning algorithm to combine the results of several other learning algorithms. Here, we regard each attention map as a channel and use the convolution operation to achieve stacking:

$$\begin{cases} \overline{s^l} = \Omega^l(S^l) \\ \Omega^l(s) = s \xrightarrow{CONV} s' \xrightarrow{sigmoid} \overline{s^l} \end{cases} \tag{8}$$

where $CONV$ is the abbreviation for the convolution layer.

As shown above, stacking can be expressed by a series of convolution layers and the final sigmoid layer, so that it can be trained directly.

### 3.3. Multi-level attention maps fusion

The fusion pipeline is presented in Fig. 5. It consists of three parts: (1) multi-level attention distributions, which is computed with the fine-grained attention. (2) Multi-resolution attention maps, which is generated from each layer of the progressive network. (3) The top-down attention fusion by cascade up-sampling.

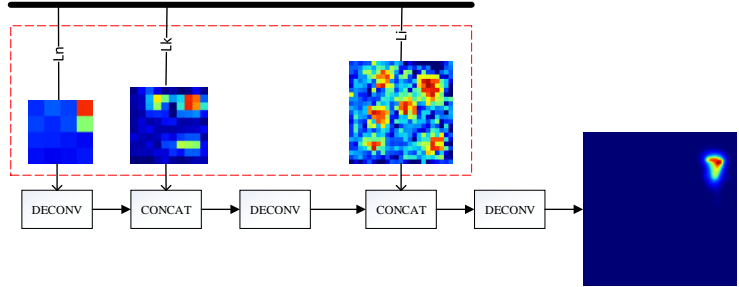

**Fig. 5.** Architecture of the learnable cascade up-sampling.

### 3.3.1. Attention maps pyramid

Due to the hierarchical property of FPAN, the attention map generated in each layer of the network results in the attention maps pyramid, whose structure is very similar to the pyramid in [35]. Maps shown within the red dashed box in Fig. 5 is an intuitive example. Specifically, an L layer of FPAN can collect a attention maps pyramid $A_{tt} = \{\overline{s^l}, \overline{s^2}, ..., \overline{s^L}\}$, where $\overline{s^L} \in R^{W^k \times H^k \times 1}$ represents the attention map generated in the k-th layer. This attention maps pyramid has the following properties:

(1) Multi-level attention distribution. The distribution of attention gradually converges to the target area, while spatial information and resolution of the attention map decrease layer by layer.
(2) The perceptual field of each map in $A_{tt}$ corresponds to each other pointwise.

The attention maps pyramid contains complementary attention information that can be fused to generate accuracy result.

### 3.3.2. Top-down cascade up-sampling

As shown in Section 3.3.1, $\overline{s^L}$ generated at the final layer of FPAN might not produce reliable information to determine the exact location of the queried object in the original image. That is, due to the loss of resolution and information, it lacks the details and the exact location information of the target object.

Considering this phenomenon, we propose to explore the attention maps pyramid $A_{tt}$ whose property (1) indicates that the maps in the pyramid can be integrated to compensate for the phenomenon described above. In particular, fusion the detail information contained in low-level maps with the accurate attention distribution in high-level maps can greatly improve the accuracy of the localization. Thus, we generate the final attention map by fusing the L maps contained in $A_{tt}$, which can be donated as $s = \Theta(A_{tt})$.

As demonstrated in FCN [37], deconvolution is a very effective method. In order to ensure that the model is end-to-end training, we use the deconvolution to achieve the up-sampling function $\Theta(\cdot)$. At the same time, taking into account the structure and property (3) of $A_{tt}$, we propose a learnable cascaded deconvolution operation which can be described by the following formula:

$$\begin{cases} \Theta(A_{tt}) = \gamma_1\left(\zeta\left(\overline{s^1}, \gamma_2\left(\zeta\left(\overline{s^1}, \ldots, \gamma_L(\overline{s^L})\right)\right)\right)\right) \\ \gamma_k(\cdot) = deconv(\cdot) \\ \zeta(x, y) = concat(x, y) \end{cases} \quad (9)$$

where the deconvolution function $\gamma_k$ performs an up-sampling operation that takes $\gamma_{k+1}$ and $\overline{s^k}$ as input, $\zeta(x, y)$ is the concatenation function that joining the two input vectors along the third dimension. The pseudo-code of the algorithm is as follows:

---

**ALGORITHM**    Top-down cascade up-sampling.

---

***Input***: Attention maps pyramid $A_{tt} = \{\overline{s^1}, \overline{s^2}, \ldots, \overline{s^L}\}$.
***Output***: Final attention map $s \in R^{W \times H}$.
**Do for t = L, L-1, …, 1**:
    **If *t == L***
        Perform a deconvolution on the top attention map $\overline{s^L}$:
        $\overline{o^L} = \gamma_L(\overline{s^L})$, $\overline{o^L} \in R^{W^{L-1} \times H^{L-1} \times 1}$
    **else**
        Combine the inputs to get an attention map with two-channels $v^t \in R^{W^l \times H^l \times 2}$:
        $v^t = \zeta(\overline{s^t}, o^{t+1})$.
        Produce the an intermediate map $\overline{o^t} \in R^{W^l \times H^l \times 1}$ by deconvolution:
        $\overline{o^t} = \gamma_t(v^t)$

    Compute the normalized by non-linear function:
    **if t == 1**:
        $s = o^t = softmax(\overline{o^t})$
    **else**:
        $o^t = sigmoid(\overline{o^t})$
**end**

---

The final softmax layer of the cascade up-sampling structure is designed as a two-classification layer which determines whether the position is affiliated with the ground truth box. This layer can be regarded as a global optimization of the attention maps. Note that the final output $s$ is the same size as the input image.

## 4. Training methods of FPAN

In this section, we describe the learning method and loss functions used to train FPAN. We design a novel loss function, which merges the information of location and target features together so as to tell network where to take the attention. In order to make the proposed unified deep model as end-to-end training, we formulate the training process as a multi-task learning (MTL) [37] that each task is treated as an independent process by sharing features. Since deep neural networks can benefit from learning with related tasks simultaneously [38], all the tasks in our model are trained together.

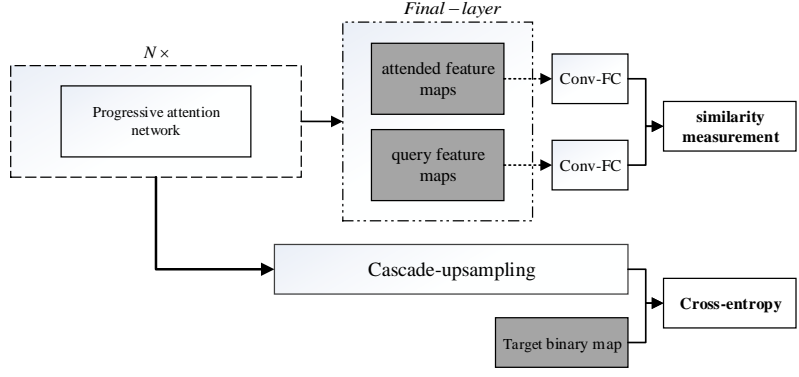

**Fig. 6.** Multi-task Learning architecture.

Inspired by the region proposal network from the segmentation-aware CNN model shown in [18] and segment loss of FCN [36], we generate a position-sensitive binary map the same size as the input image, with the background and the ground truth box as class 0 and class 1, respectively. The cross-entropy is used to identify the difference between this map and the final attention map. We also compare the similarity between the attended features and the template features at the end of the network to drive the network to locate the target region.

In general, training a unified deep model that incorporates multi-level independent supervised information is non-trivial, because different level information has various learning difficulties and convergence rates [37]. However, the two tasks we define are mutually reinforcing. Since image and query share the feature extraction layer, the high similarity between the attended features and the template features indicates that the regions unrelated to the query are completely suppressed in attention distribution which also leads to accurate localization. vice versa. That is to say, such a mechanism enables our learning process to learn how the tasks should interact with each other, which is crucial for multi-task learning [37].

Based on the above principle, the proposed loss function described above are shown in Eq. (10) and Fig. 6.

Given N training samples in a mini-batch, the goal of our model is trying to minimize

$$\min_{W=\{W_1,W_2\}} \sum_{i=1}^{N} L_{seg}(\tilde{y}^{(i)}, s; W_1, W_2) - L_{sim}\left(\left(f_q^l\right)^{(i)}, f\left((x^{(i)}, q^{(i)}); W_1\right)\right) + R(W) \quad (10)$$

where $\left(x^{(i)}, q^{(i)}, \tilde{y}^{(i)}\right)$ is a training sample, $q^{(i)}$ is the queried object, $x^{(i)}$ is the image needs to be searched and $\tilde{y}^{(i)}$ is the target binary map. W represents all the weights of our model while $f\left((x^{(i)}, q^{(i)}); W_1\right)$ is the forward procedure that outputs the attended feature map of $x^{(i)}$. $L_{seg}(\cdot)$ is cross entropy loss. $L_{sim}(\cdot)$ is the similarity measure. Additionally, a regularization term $R(W)$ is added to the target function.

For the loss function $L_{seg}(\cdot)$, which is box segmentation loss, we define the following formula:

$$L_{seg}(\tilde{y}, s) = \text{cross} - \text{entropy}(\tilde{y}, s), \ \tilde{y}, s \in R^{W \times H} \quad (11)$$

Here, the target binary map $\tilde{y}$ is defined as follows:

$$\tilde{y}(i,j) = \begin{cases} 1, & i \in [x_{tl}, x_{br}], j \in [y_{tl}, y_{br}] \\ 0, & else \end{cases} \quad (12)$$

where $(x_{tl}, y_{tl})$ and $(x_{br}, y_{br})$ denote the top-left and bottom-right coordinates of the ground-truth, respectively.

The loss function $L_{sim}(\cdot)$ seeks to push the net to locate the target features while also obtain aligned representations of the query template and the target in the image which can be achieved by the Eq. (13).

$$\begin{cases} f_{attn}^L = s^L \odot f^L \\ v_{attn}^L = conv\_FC(f_{attn}^L) \\ v_q^L = conv\_FC(f_q^L) \\ C(v_{attn}^L, v_q^L) = \cos(v_{attn}^L, v_q^L) \end{cases} \quad (13)$$

where cos is the cosine of the angle between the two representation vectors. Note that $conv\_FC$ is a fully-connected layer implemented with the convolution operation, as shown in FCN [36].

**5. Experiments**

In this section, we empirically analysis our proposed approach on two typical localization tasks and present the comparisons with popular methods on the synthetic custom datasets and benchmark datasets, respectively. In the synthetic datasets, we locate the queried digit on the given image. Then we perform experiments on object tracking to fully test the performance of the algorithm. Our algorithm is implemented in Tensorflow-1.4 with Python wrapper and runs at around 40 fps with eight cores of 3.4GHz Intel Core i7-6700 and an NVIDIA Tesla K20m GPU.

5.1. Digital retrieval

The proposal of this experiment is to locate the target object closest to the provided template in the image. The details of the experiment are as follows:

***Datasets*** Based on the dataset introduced in [29] which is a very classical dataset to test the performance of attention-based detection methods, we synthesize the data set called MNIST-Q and MNIST-RQ. Instead of getting the color of the input query, we change the task to a template-based digital localization task.

**MNIST-Q** This dataset has 20000 images and 9 colored digits, with at least five digits per image. For each image, one of the colored digit patch with size $28 \times 28$ is used as a query with the corresponding ground-truth box used as the label. For the contents of each picture in the MNIST, it consists of a black background, a series of colored digits and a large number of noise pixels, shown in Fig.7 (a).

**MNIST-RQ** This dataset consists of 10000 images that is essentially similar to MNIST-Q with only more complex scenes. That is, the background of the image comes from the COCO [39] and Sun dataset [40]. An example of this dataset is shown in Fig. 5(b).

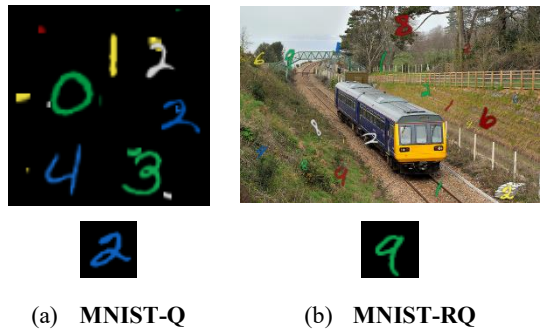

(a)　**MNIST-Q**　　　(b)　**MNIST-RQ**

**Fig. 7.** Examples of the two datasets where the first row is the image needed to be searched, the second row is the template of the queried object.

***Baseline Algorithms*** Since this is a typical template matching problem, we use the classic template matching algorithms as the benchmark. On the one hand, we use the sliding window algorithm provided in OpenCV which uses the correlation and correlation coefficient as the measure of the similarity. On the other hand, we use the regions proposal method combined with Hog feature [42]. Specifically, we use the selective search method provided in RCNN [42] to generated a series of boxes, then Hog features will be extracted from these boxes and template. Finally, the square difference measure method is used to select the similar ones. For both methods, the color histogram is used to determine the target box. The above two methods are abbreviated as OPV-TM and PHH, respectively. In addition, the FPAN without cascade up-sampling layer is donated as FPAN_NO-DE and FPAN-SS is the FPAN without fine-grained attention used.

***Evaluation metrics*** To evaluate different algorithms, average localization precision (ALP), average IOU of the success located digit (AIOU) and average time spent by each query (T) are adopted as the standard quantitative measurement. In addition, the IOU-precision curve is shown to further analysis algorithms.

### 5.1.1. Implementation Details

***Feature extractor*** A four layer of CNN is used as the basic feature extractor, with the structure and the setting shown in Table 2. In each level the structure, the feature extractor is shared by the image and the query patch.

***Fine-grained attention*** The structure and setting of the Fine-grained attention model are shown in Table 1. Image-FP is the image feature maps and Patch-FP is the template feature maps which will forward the MFE discussed in subsection 3.1.1. The generated attention maps are AM1 ~ AM4. Finally, a convolution of size 3x3 is used to generate the final attention map Fusion-AM. Attn-Image-FP is obtained by multiplying Image-FP and Fusion-AM by elements, serves as the input image feature of the next layer.

***Training*** The attention network presented in this paper is a unified, full-convolution end-to-end network so that it can be trained directly. The loss function is described in Section4. In training, all parameters are initialized with Xavier [43] while the RMSprop [44] is used as the optimizer. The initial value of the learning rate is 0.03 and varies according to the exponential decrease. In detail, we randomly select a digit contained in the image as the target object and the corresponding digit template as the query.

**Table 1**

Structure of the fine-grained attention.

| Fine-grained attention | | | | |
|---|---|---|---|---|
| Image-FP | | | | Patch-FP |
| CSCONV: 3x3, 1x1 | CSCONV: 3x3, 1x1 | CSCONV: 5x5, 1x1 | POOL: 2x2, 1x1 | MFE |
| ↓ | CSCONV: 3x3, 1x1 | ↓ | ↓ | ↓ |
| | + | | | |
| AM1 | AM2 | AM3 | | AM4 |
| Conv: 3x3, 1x1 | | | | |
| Fusion-AM | | | | |
| Attn-Image-FP | | | | |

**Table 2**

Structure of the FPAN model.

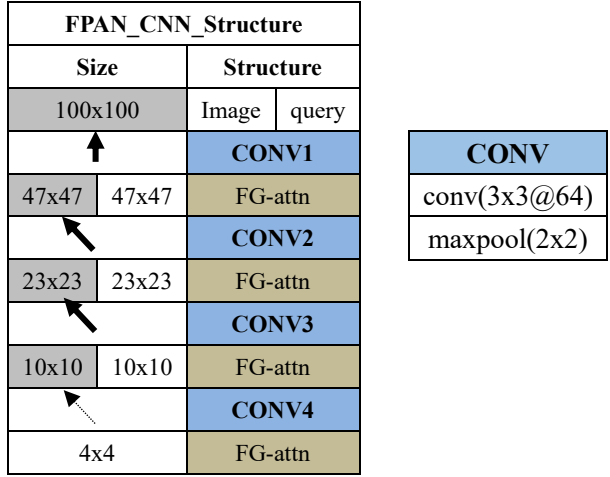

### 5.1.2 Experimental results

**Table 3**

The results of AIOU, Precious (%) and Time on MNIST-Q when the IOU threshold for a success retrieval is 0.5.

| Algorithms | | AIOU | ALP | T |
|---|---|---|---|---|
| **Ours** | FPAN | 0.88 | 90.5 | 0.08 |
| | FPAN_SS | 0.845 | 87 | 0.03 |
| | FPAN_NO-De | 0.788 | 86.5 | 0.06 |
| **OPV_MT** | TM_CCOEFF_NORMED | 0.637 | 41.2 | 0.094 |
| | TM_CCORR_NORMED | 0.662 | 44.5 | 0.48 |
| **PHH** | Proposal+Hog_SVM+Hist | 0.744 | 85.8 | 0.28 |

**Table 4**

The results of AIOU, Precious (%) and Time on MNIST-RQ when the IOU threshold for a success retrieval is 0.5.

| Algorithms | | AIOU | ALP | T |
|---|---|---|---|---|
| **Ours** | FPAN | 0.71 | 70 | 0.8 |
| | FPAN-SS | 0.695 | 54.7 | 0.4 |
| | FPAN-NO-De | 0.626 | 64.8 | 0.7 |
| **OPV_TM** | TM_CCOEFF_NORMED | 0.62 | 11.8 | 21 |
| | TM_CCORR_NORMED | 0.623 | 13.4 | 49 |
| **PHH** | Proposal+Hog_SVM+Hist | 0.565 | 46 | 2.2 |

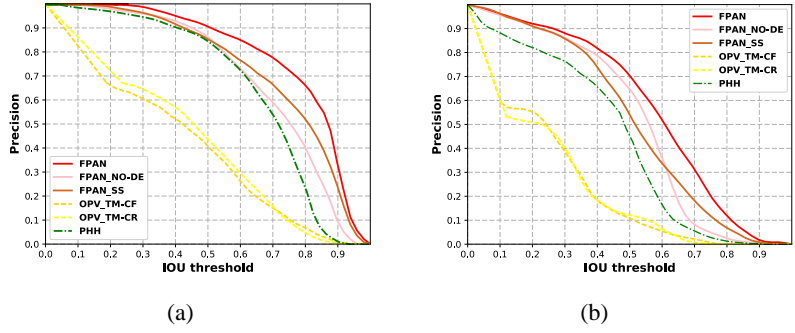

**Fig. 8.** IOU-precision curve.

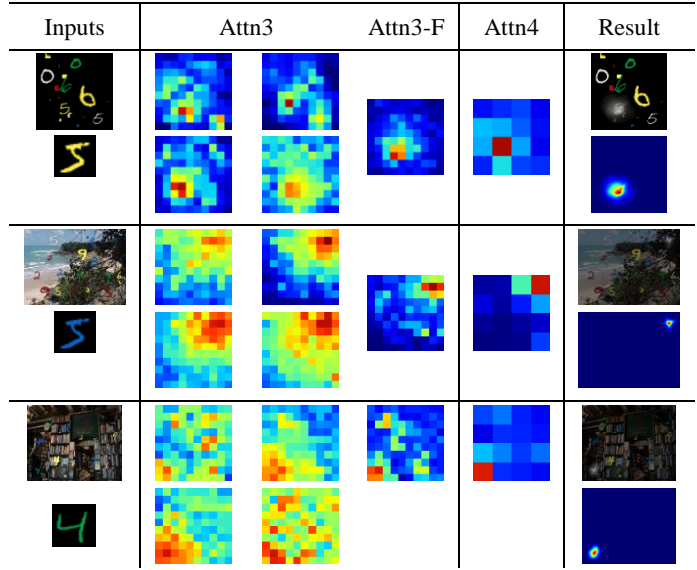

**Fig. 9.** Visualization of the results with all the attention maps rendered by the standard heatmap.

We compare FPAN with all other baseline algorithms on MNIST-Q and MNIST-RQ with the evaluation schema described in 5.1. The average results for all test data are summarized in Table 3 and Table 4, and the IOU-precision curve is displayed in Fig. 8. In order to facilitate the in-depth analysis of the algorithm, some typical renderings are shown in Fig. 9.

As shown in Table 3 and Table 4, FPAN performs the best results with the minimum time used. In the following, we analyze the experimental results and their deep reasons from the aspects of quantitative, qualitative and efficiency.

*Quantitative Comparison* The quantitative comparisons in terms of the AIOU, ALP and T are shown in Table 3 and Table 4 and Fig. 8.

From the quantitative results, we can conclude that our approach (FPAN) outperforms others benchmark algorithms on all the metrics with large margins. Specifically, it improves by 0.14, 4.7% and 0.09, 24% over the best-performing algorithm according to the AIOU, ALP scores on MNIST-Q, MNIST-RQ, respectively. In addition, from the IOU-precision curve in Fig. 8, FPAN can achieve best localization precision under any IOU threshold. Especially in the challenging cases where the background is complex and cluttered in MNIST-RQ dataset, from the results in Table 4 and Fig. 8 (b), our algorithm is 47% higher than the average precision of other algorithms.

In the internal comparison of FPAN, FPAN_SS and FPAN_NO-DE, the impact of the fine-grained attention module on the model is mainly reflected in AIOU, which is lower than FPAN 0.35, 15.3 on

MNIST-Q, MNIST-RQ, respectively, while the cascade up-sampling has a great influence on ALP which can improve the performance of the model 0.1 and 0.84 on MNIST-Q, MNIST-RQ, respectively.

***Qualitative Analysis*** We also provide a qualitative result of FPAN in Fig. 9, which is the visual outputs of the network. We mainly analyze the high-level attention results and provide a demonstration of the final localization results. In the fine-grained attention structure of the third layer, as shown in the second column of Fig. 9, we obtain four attention maps. It can be seen that the fused Attn3-F suppresses irrelevant information and focuses on the target area. As can be seen from the attention map Attn4-F in the fourth layer, as shown in the fourth column of Fig. 9, model focuses attention accurately on the target location. By comparing Attn3-F and Attn4-F, we can see that the distribution of attention is more concentrated and more accurate as the layers deepen, that is, the attention gradually shrinks to the target location layer by layer. It is also worth noting that our algorithm uses cascading up-sampling to obtain a tight package for the target object in the original image. As can be seen from the fourth column of the resulting maps, our method almost successfully highlights the entire target object.

***Efficiency*** Since the whole network is an end-to-end unified network, the algorithm can be run directly with a one-stage computation. At the same time, the FPAN is a fully convolution neural network structure where convolution operations can be accelerated by GPUs. We utilize the multi-core (2TSL in our experiment) to accelerate the computation. From the column of T in Table 3 and Table 4, we can see, our approach can run 3-20 times faster than most existing methods.

## 5.2. Tracking performance

Tracking a target object is a crucial task that involves separating the target from the background and precisely locating the target object which is a huge challenge for the localization network. To test the performance of the proposed FPAN dealing with the general tracking problem, we use several commonly used datasets and evaluation metrics as benchmarks.

***Benchmark Datasets*** The experiments are conducted on three standard benchmarks: OTB-2013 [45], OTB-2015 [46] and VOT-2016 [47]. The first two datasets are popular tracking benchmark that contain 50 and 100 fully annotated videos with substantial variations. In the VOT-2016 dataset, there are 60 challenging videos from a set of more than 300 videos.

***Evaluation Metrics*** The evaluation mainly follows the standard metrics from the benchmarks. For the OTB-2013 and OTB-2015, we use the one-pass evaluation (OPE) with precision and success plot metrics. The precision metric measures the rate of frame locations within a certain threshold distance from those of the ground truth. The threshold distance is set as 20 for all the trackers. The success plots metric measures the overlap ratio between predicted and ground truth bounding boxes. For the VOT-2016 dataset, the performance is measured in terms of expected average overlap (EAO), accuracy (A), robustness (R), accuracy ranks (Ar), and robustness ranks (Rr).

***Benchmark Algorithms*** We compare our FPAN tracker with several classic tracking algorithms, including MDNet [48], C-COT [49], Deep SRDCF [50], HCFT [51], MUSTer [52], SRDCF [53], CNN-SVM [54], KCF [55], MEEM [56], Struck [57], FCNT [58], TCNN [59], ColorKCF [60], ACT [61], DNT [62], Staple [63].

As the experimental method in Section 5.1, two variations of FPAN are implemented to verify the contribution of each component in our algorithm, including FPAN_NO-DE and FPAN-SS.

### 5.2.1. Implementation details

As discussed in the Section 5.1, the structure of our algorithm mainly consists of three parts. In addition, the FPAN tracker also requires additional design and elaboration which are mainly described next.

*Feature Extractor* The design details of the main feature extraction network are shown in Table5. The basic network structure is based on the pre-trained VGG19 [64] with first 16 layers used as feature extractor.

*Fine-grained Attention* For the structure of the fine-grained attention, we use the same setting as 5.1 shown in Table 2.

*Training* For training of FPAN in OTB, we use training sequences collected from VOT2016, excluding the videos included in OTB. vice versa. For details, we obtain the target patch from the first frame. Then, to locate the target object in each frame $f_i$, we use the result $s_{i-1}$ of the previous frame $f_{i-1}$ as the query object and crop the box two times bigger than $s_{i-1}$ from the center location of $s_{i-1}$ in $f_i$ as the search area. We regard the target object localization in two consecutive frames as a sample. It should be noted that when creating the training set, we store the samples from all videos out of order, that is, there is no time and the content correlation between two adjacent samples. In the training stage, we iteratively apply the Adam optimizer [65] with a learning rate of 4e-6 to update the model, until the loss is below the given threshold 0.05 to prevent the over-fitting. Meanwhile, all the parameters except the feature extractor are randomly initialized following zero-mean Gaussian distribution.

*Online Learning* The online learning scheme is straightforward in our tracker. Since our model does not rely on the continuity of content in the video frame sequence but instead directly locates the queried object for each frame, no complicated online update technique is required.

**Table 5**

Structure of the FPAN model

| FPAN_CNN_Structure | | |
|---|---|---|
| Size | | Structure |
| $W \times H$ | | Image / query |
| | | **VGG-19: L1-L2** |
| $W/2 \times H/2$ | $W/2 \times H/2$ | FG-attn |
| | | **VGG-19: L3-L15** |
| $W/16 \times H/16$ | $W/16 \times H/16$ | FG-attn |
| | | **VGG-19: L16** |
| $W/32 \times H/32$ | | FG-attn |

5.2.2. Experimental results

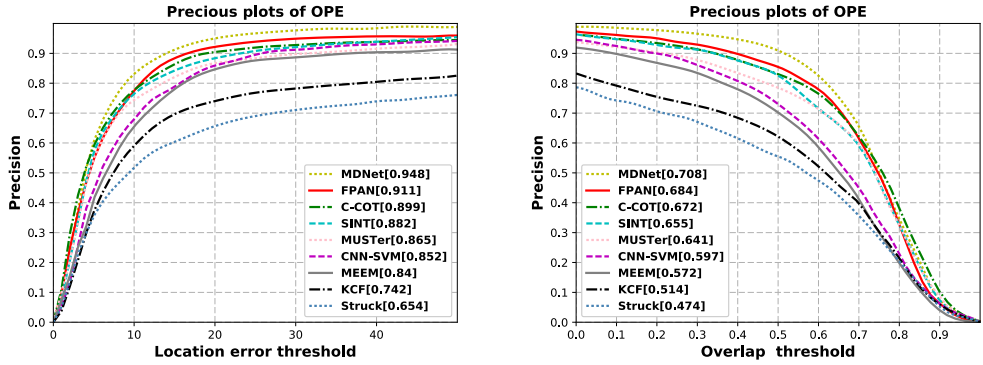

(a) OTB50 results

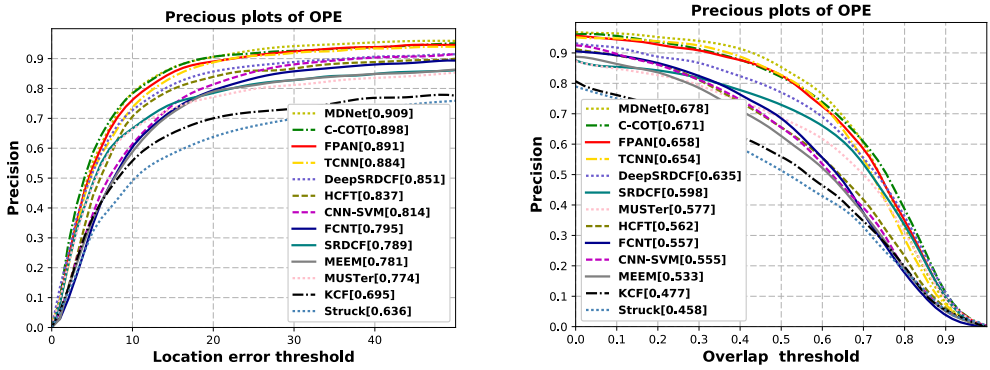

(b) OTB100 results

**Fig. 10.** Precision and success curves of OTB50 and OTB100. The numbers of precisions at 20 pixels for precision and the area-under-curve for success plots are in in descending order.

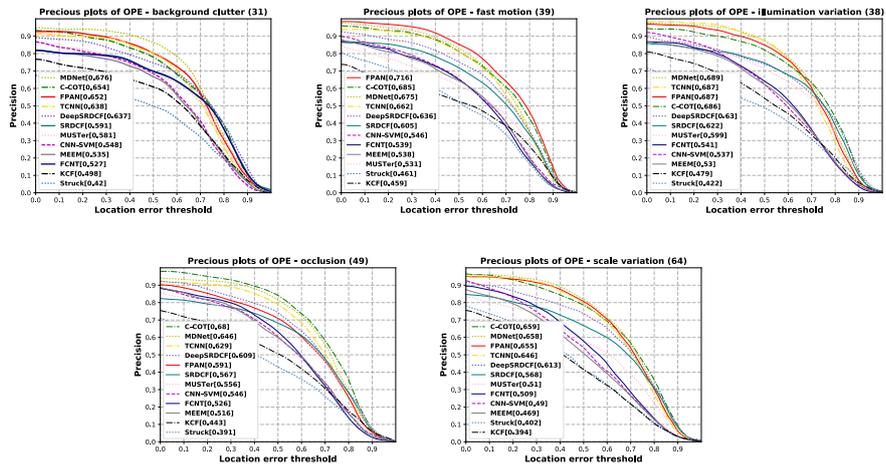

**Fig. 11.** The results in OTB100 for 5 typical situations: background clutter, fast motion, illumination variation, occlusion and scale variation.

**Table 6**

Comparison with the several trackers on the VOT 2016 dataset. The results are presented in terms of expected average overlap (EAO), accuracy (A), robustness (R), accuracy ranks (Ar), and robustness ranks (Rr).

| Tracker | EAO | A | R | Ar | Rr |
|---|---|---|---|---|---|
| FPAN | 0.296 | 0.537 | 0.266 | 3.000 | 2.000 |
| FPAN_SS | 0.268 | 0.503 | 0.345 | 15.000 | 5.000 |
| FPAN_NO-DE | 0.231 | 0.486 | 0.423 | 9.000 | 14.000 |
| C-COT | 0.331 | 0.539 | 0.238 | 11.000 | 2.000 |
| TCNN | 0.325 | 0.554 | 0.268 | 1.000 | 4.000 |
| Staple | 0.295 | 0.544 | 0.378 | 8.000 | 14.000 |
| DNT | 0.278 | 0.515 | 0.329 | 21.000 | 7.000 |
| MDNet | 0.257 | 0.541 | 0.337 | 11.000 | 7.000 |
| SRDCF | 0.247 | 0.535 | 0.419 | 11.000 | 18.000 |
| ColorKCF | 0.226 | 0.503 | 0.443 | 21.000 | 19.000 |
| KCF | 0.192 | 0.489 | 0.569 | 37.000 | 37.000 |
| ACT | 0.173 | 0.446 | 0.662 | 49.000 | 47.000 |
| FCT | 0.141 | 0.395 | 0.788 | 64.000 | 56.000 |

As shown in 5.2.1, we evaluate the proposed FPAN tracker on two public benchmark datasets, Object Tracking Benchmark(OTB) and VOT2016 and compare with all other benchmark algorithms in the listed metric. In this section, we show and analyze the experimental results.

***Evaluation on OTB*** we present the results on both OTB-100 and OTB50. Fig. 10 shows the evaluation results on OTB-100 and OTB-50. It can be seen that among all the trackers, our FPAN tracker outperforms most of the trackers on both datasets. Specifically, our algorithms work well with both slight and strict thresholds, indicating that FPAN can find tight bounding box without losing the target easily. Besides, the performances of all the variations methods are not as good as the full FPAN which means each component in FPAN tracker is essential.

In addition, we further analyze the performance under various challenge attributes (e.g., background clutter, fast motion and occlusion etc.). The one pass evaluation on the AUC score under five main video attributes are shown in Fig. 11 which demonstrates that our tracker can effectively handle most situations with high performance. The challenge situation such as background clutter often require the model can understand high-level semantic information, our model has a better effect on these tasks. It is mainly because that the fine-grained attention structure can capture enough local information for effective distribution of spatial attention values. Meanwhile, the reason why our algorithm works well in the scale variation and fast motion video sequences is mainly because of cascading up-sampled structure which can repair and adjust the attention distribution so as to provide the model with robustness to scale variation and position change. However, in occlusion sequences, our model did not achieve as good results as the others. This can be attributed to the fact that the distribution of high-level attention is largely based on the lower-level attention distributions, which means that it is difficult to repair deviations in the progressive attentive process. Therefore, in the future, we will consider the forgotten-memory structure to adjust the network's over-dependence on low-level attention maps.

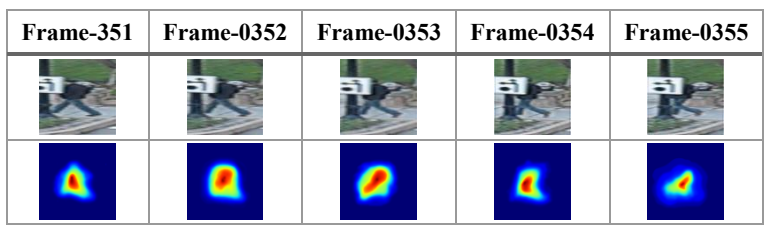

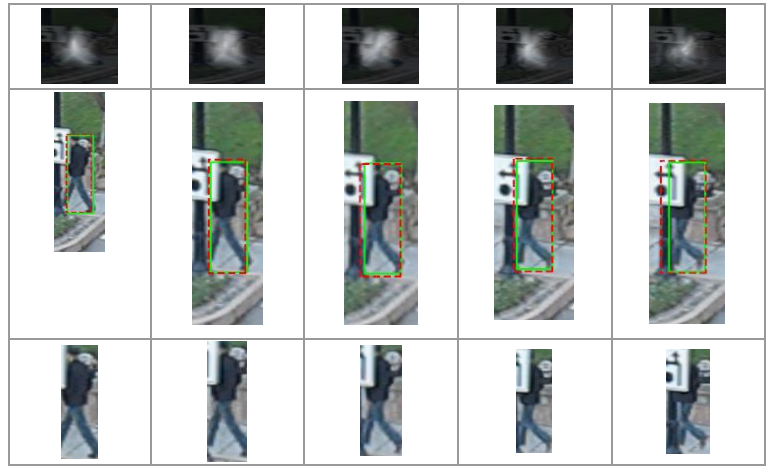

(a)

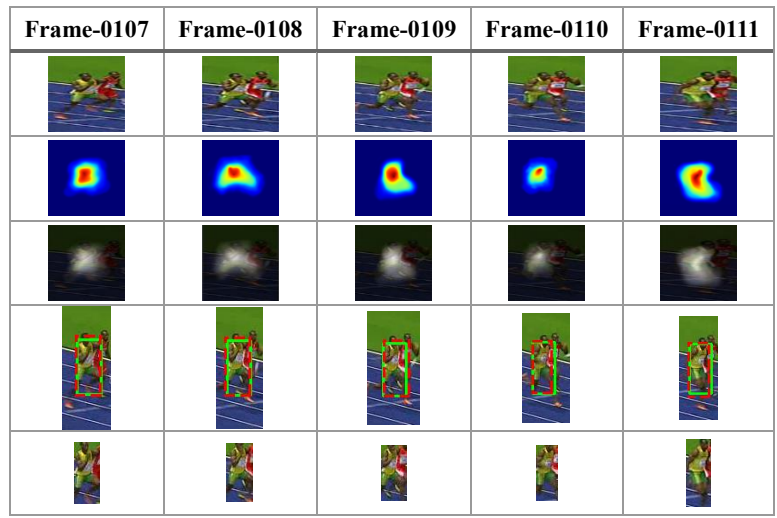

(b)

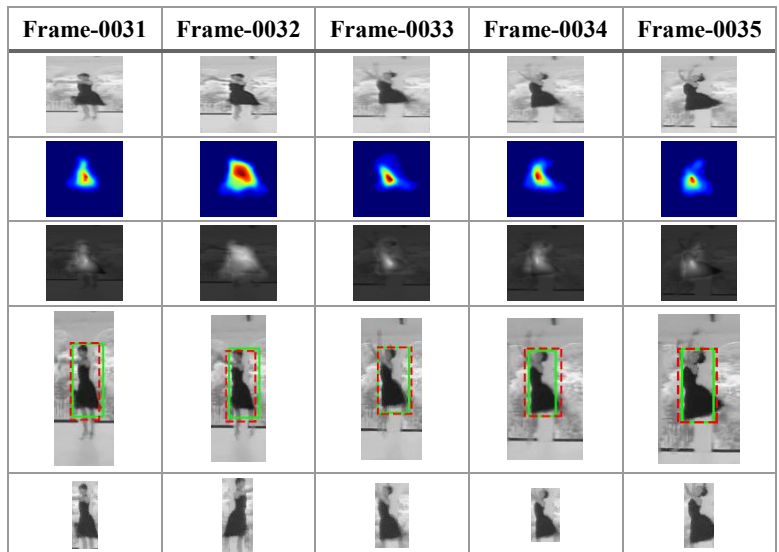

(b)

**Fig. 12.** The visualization results on consecutive sequences. The first row is the label of the frame. The second row is the cropped image needed to be searched. The third row is the final attention map while the correspond brightness comparison of the image is shown in fourth row. The localization result is shown in fifth row, where the red box is the result of FPAN and green one is the ground truth. The last row is the query cropped from current frame used as a query for the next frame.

*Evaluation on VOT* To better evaluate the algorithm, we perform the experiment described in 5.2.1 on the VOT2016 dataset. Table 6 shows the results from our FPAN tracker and other classic trackers. For tracker whose EAO values exceed the state-of-the-art bound 0.251 are all considered as state-of-the-art trackers. As can be seen, our tracker is the state-of-the-art as other classic algorithms. Among these methods, FPAN also achieves the best results under the most metrics.

In addition, as shown in Fig.12, we visualize several tracking results on some video sequences. In order to better reflect the idea that FPAN uses adjacent frames to track, we mainly show the results of continuous frames. Overall, FPAN can accurately locates the target object with a tight package while keeping up with the target.

## 6. Conclusion

In this paper, we study the data retrieval problem of automatically and accurately locating the queried object in the image. We propose a unified framework that uses fine-grained attention mechanism to generate effective and stable attention distributions while gradually suppressing the uninterested areas in the progressive attention network so as to focus attention on the target object. Furthermore, the accurate localization of the queried object is achieved by the cascade up-sampling structure. We formulate a multi-task learning schema that jointly optimizes detection and location tasks, thus training the proposed network to intelligently and efficiently perform the data retrieval task. In the query-based digital localization experiment, it achieves optimal results on both the performance and efficiency. The proposed algorithm overcomes the shortcomings of traditional query-based localization approaches, which is time-consuming and inefficient. We also extend our model to the object tracking tasks in which the proposed method provides an effective solution to the retrieval tasks with high speed and favorable precision. This suggests that our framework can be suitable for practical applications that related to rapid data retrieval. An interesting direction for future research is the residual connection of attention maps in the progressive attention network.

## References

[1] S. Korman, D. reichman, G. Tsur, S. Avidan, Fast-Match: Fast Affine Template Matching, in: International Journal of Computer Vision, 2017, Vol. 121, Issue. 1, pp. 111–125.

[2] P. Swaroop, N. Sharma, An Overview of Various Template Matching Methodologies in Image Processing, in: International Journal of Computer Applications 153(10):8-14, 2016.

[3] A. Ghodrati, A. Diba, M. Pedersoli, T. Tuytelaars, Luc Van Gool, DeepProposal: Hunting Objects by Cascading Deep Convolutional Layers, in: IEEE International Conference on Computer Vision (ICCV), 2015, vol. 00, pp. 2578-2586.

[4] I., Laurent, C. Koch, E. Niebur, A model of saliency-based visual attention for rapid scene analysis, in: IEEE Transactions on Pattern Analysis & Machine Intelligence, 1998, pp. 1254-1259.

[5] Desimone, Robert, J. Duncan, Neural mechanisms of selective visual attention, in: Annual review of neuroscience, 1995, pp. 193-22.

[6] V. Mnih, N. Heess, A. Graves, et al, Recurrent models of visual attention, in: International Conference on Neural Information Processing Systems, 2014, Vol. 2, pp. 2204-2212.

[7] K. Xu, J. Ba, R. Kiros, Show, Attend and Tell: Neural Image Caption Generation with Visual Attention, in: Computer Science, 2015, pp. 2048-2057.

[8] S. E. Kahou, V. Michalski, R. Memisevic, RATM: Recurrent Attentive Tracking Model, in: Computer Science, 2015.

[9] M. Biparva, J. K. Tsotsos, STNet: Selective Tuning of Convolutional Networks for Object Localization, in: arXiv:1708.06418,

2017.

[10] J.R.R. Uijlings, K.E.A. van de Sande, T. Gevers, A.W.M. Smeulders, Selective Search for Object Recognition, in: Proc. International Journal of Computer Vision, 2013, Vol. 104, Issue. 2, pp. 1185–1192.

[11] P. Rantalankila, J. Kannala, E. Rahtu, Generating object segmentation proposals using global and local search, in: Computer Vision & Pattern Recognition (CVPR), 2014, pp: 2417-2424.

[12] M. M. Cheng, Z. Zhang, W.-Y. Lin, and P. H. S. Torr, BING: Binarized normed gradients for objectness estimation at 300fps, in: Computer Vision & Pattern Recognition (CVPR), 2014, pp: 3286-3293.

[13] C. Zitnick, P. Dollár, Edge boxes: Locating object proposals from edges, in: European Conference on Computer Vision (ECCV), 2014, pp. 391-405.

[14] B. Alexe, V. Petrescu, V. Ferrari, Exploiting spatial overlap to efficiently compute appearance distances between image windows, in: Advances in Neural Information Processing Systems (NIPS), 2011, pp. 2735-2743.

[15] J. Yu, J. Amores, N. Sebe, P. Radeva, Q. Tian, Distance Learning for Similarity Estimation, in: IEEE Transactions on Pattern Analysis and Machine Intelligence, 2008, Vol. 30, Issue. 3, pp: 451-462.

[16] S. Ren, K. He, R. Girshick, Faster R-CNN: Towards real-time object detection with region proposal networks, in: IEEE Transactions on Pattern Analysis and Machine Intelligence, 2017, Vol. 39, Issue. 6, pp. 1137-1149.

[17] J. Dai, Y. Li, K, He, J. Sun, R-FCN: Object Detection via Region-based Fully Convolutional Networks, in: arXiv:1605.06409.

[18] S. Gidaris, N. Komodakis, Object detection via a multi-region & semantic segmentation-aware CNN model, in: Computer Vision (ICCV), 2016, pp: 1134-1142.

[19] S. Gidaris, N. Komodakis, LocNet: Improving Localization Accuracy for Object Detection, in: Computer Vision and Pattern Recognition (CVPR), 2016, pp: 789-798.

[20] X. He, Y. Peng, J. Zhao, Fine-grained Discriminative Localization via Saliency guided Faster RCNN, in: arXiv:1709.08295..

[21] X. Zhang, H. Xiong, W. Lin, Q. Tian, Ensemble of Part Detectors for Simultaneous Classification and Localization, in: arXiv:1705.10034.

[22] B. Zhou, A. Khosla, A. Lapedriza, et. al., Learning Deep Features for Discriminative Localization, in: Computer Vision and Pattern Recognition (CVPR), 2016, pp. 2921-2929.

[23] C. Y. Fu, W. Liu, A. Ranga, A. Tyagi, A. C. Berg, DSSD : Deconvolutional Single Shot Detector, in: arXiv:1701.06659,

[24] A. Torralba, A. Oliva, M. S. Castelhano, et al, Contextual guidance of eye movements and attention in real-world scenes: the role of global features in object search, in: Psychological Review, 2006, pp. 766-786.

[25] M. Wischnewski, A. Belardinelli, Schneider W X, et al, Where to Look Next? Combining Static and Dynamic Proto-objects in a TVA-based Model of Visual Attention, in: Cognitive Computation, 2010, pp. 326-343.

[26] M. Denil, L. Bazzani, H. Larochelle, et al, Learning Where to Attend with Deep Architectures for Image Tracking, in: Neural Computation, 2012, pp. 2151-2184.

[27] S. Sharma, R. Kiros, R. Salakhutdinov, Action Recognition using Visual Attention, in: International Conference on Learning Representations (ICLR), 2016.

[28] D. Yoo, S. Park, J. Y. Lee, et al, AttentionNet: Aggregating Weak Directions for Accurate Object Detection, in: IEEE International Conference on Computer Vision (ICCV), 2015, pp. 2659-2667.

[29] P. H. Seo, Z. Lin, S. Cohen, et al, Progressive Attention Networks for Visual Attribute Prediction, in: International Conference on Learning Representations (ICLR), 2016.

[30] F. Chollet, Xception, Deep Learning with Depthwise Separable Convolutions, in: arXiv:1610.02357.

[31] MD. Zeiler, R. Fergus, Visualizing and Understanding Convolutional Networks, in: Springer International Publishing, 2014, pp. 818-833.

[32] C. Szegedy, V. Vanhoucke, S. Loffe, J. Shlens, Z. Wojna, Rethinking the Inception Architecture for Computer Vision, in: Computer Vision and Pattern Recognition (CVPR), 2016, pp. 2818-2826.

[33] R. Polikar, Ensemble based systems in decision making, in: IEEE Circuits and Systems Magazine, 2006, pp. 21-24.

[34] D. Wolpert, Stacked Generalization., in: Neural Networks, 1992, vol. 5, issue 2, pp. 241-259

[35] T. Harada, Y. Ushiku, Y. Yamashita, Y. Kuniyoshi, Discriminative Spatial Pyramid, in: Computer Vision and Pattern Recognition (CVPR), 2011, pp. 1617-1624.

[36] J. Long, E. Shelhamer, T. Darrell, Fully Convolutional Networks forSemantic Segmentation,    in: IEEE Computer Vision and Pattern Recognition, 2015, pp. 3431-3440.

[37] A. Evgeniou, M. Pontil, Multi-task feature learning, in: Neural Information Processing Systems, 2006, Vol. 19, issue 3, pp. 41-48.

[38] Ruder. S, An Overview of Multi-Task Learning in Deep Neural Networks, in: International Journal of Data Warehousing and Mining, 2007.

[39] TY. Lin, M. Maire, et. al, Microsoft COCO: Common Objects in Context, in: European Conference on Computer Vision (ECCV), 2014, pp. 740-755

[40] Xiao, Jianxiong, Ehinger, Krista A, Hays, James, Torralba, Antonio, and Oliva, Aude, Sun database: Exploring a large collection of scene categories, in: International Journal of Computer Vision (IJCV), 2014, pp. 1–20.

[41] R. Girshick, J. Donahue, T. Darrell, J. Malik, Rich feature hierarchies for Accurate Object Detection and Segmentation, in: Computer Vision and Pattern Recognition (CVPR), 2014, pp: 580-587.

[42] N. Dalal, B. Triggs, et. al, Object Detection using Histograms of Oriented Gradients, in: European Conference on Computer Vision (ECCV), 2013.

[43] X. Glorot, Y. Bengio, Understanding the difficulty of training deep feedforward neural networks, PMLR, 2010, vol. 9, pp. 249-256.

[44] Tieleman, T. Hinton, Geoffrey, Lecture 6.5-rmsprop: Divide the gradient by a running

average of its recent magnitude, in: COURSERA: Neural Networks for Machine Learning, 2012.

[45] Y. Wu, J. Lim, M. H. Yang, Online object tracking: A benchmark, in: Proceedings of the IEEE Computer Society Conference on Computer Vision and Pattern Recognition, 2013, pp. 2411-2418.

[46] Y. Wu, J. Lim, M.-H. Yang, Object tracking benchmark, in: IEEE Transactions on Pattern Analysis and Machine Intelligence (TPAMI), 2015, vol. 37, issue 9, pp. 1834-1848.

[47] M. Kristan, et al, The visual object tracking vot2016 challenge results, in: European Conference on Computer Vision, 2016, pp. 1-27.

[48] H. Nam, B. Han, Learning multi-domain convolutional neural networks for visual tracking, in: VOT (Visual Object Tracking) Challenge, (2015).

[49] M. Danelljan, A. Robinson, F. S. Khan, M. Felsberg, Beyond correlation filters: Learning continuous convolution operators for visual tracking, in: Lecture Notes in Computer Science, 2016, vol. 9909, pp. 472-488.

[50] M. Danelljan, G. Häger, F. Khan, M. Felsberg. Convolutional features for correlation filter based visual tracking. In: Proceedings of the IEEE International Conference on Computer Vision, Vol. 2016–February, pp. 621–629.

[51] C. Ma, J.-B. Huang, X. Yang, M.-H. Yang, Hierarchical convolutional features for visual tracking, in: International Conference on Computer Vision (ICCV), 2015, vol. 2015, pp. 3074-3082.

[52] Z. Hong, Z. Chen, C. Wang, X. Mei, D. Prokhorov, D. Tao, MUlti-Store Tracker (MUSTer): a cognitive psychology inspired approach to object tracking, in: Proceedings of the IEEE Computer Society Conference on Computer Vision and Pattern Recognition, 2015, Vol. 07-12-June-2015, pp. 749–758.

[53] M. Danelljan, G. Häger, F. Khan, M. Felsberg, Learning spatially regularized correlation filters for visual tracking. in: Proceedings of the IEEE International Conference on Computer Vision (ICCV), 2015, pp. 4310–4318.

[54] S. Hong, T. You, S. K. wak, B. Han, Online tracking by learning discriminative saliency map with convolutional neural network, in: Proceedings of the 32nd International Conference on International Conference on Machine Learning, 2015, Vol. 37, pp. 597-606.

[55] J. F. Henriques, R. Caseiro, P. Martins, and J. Batista, Exploiting the circulant structure of tracking-bydetection with kernels,

in: Lecture Notes in Computer Science, 2015, Vol. 7575 LNCS, pp. 702–715.

[56] J. Zhang, S. Ma, S. Sclaroff, Meem: robust tracking via multiple experts using entropy minimization, In: Lecture Notes in Computer Science, 2014, Vol. 8694 LNCS, issue 6, pp. 188-203.

[57] S. Hare, A. Saffari, P. H. Torr, Struck: Structured output tracking with kernels, in: IEEE Transactions on Pattern Analysis and Machine Intelligence, 2016, vol. 38, issue 10, pp. 2096-2109.

[58] L. Wang, W. Ouyang, X. Wang, H. Lu, Visual tracking with fully convolutional networks, IEEE International Conference on Computer Vision (ICCV), 2015, pp. 3119-3127.

[59] H. Nam, M. Baek, B. Han, Modeling and propagating cnns in a tree structure for visual tracking, in: arXiv:1608.07242, 2016.

[61] M. Felsberg, Enhanced distribution field tracking using channel representations, in: Proceedings of the IEEE International Conference on Computer Vision, 2013, pp. 121-128

[62] Z. Z. Chi, et. al, Dual Deep Network for Visual Tracking, in: IEEE Transactions on Image Processing, 2017, Vol. 26, Issue. 4.

[63] L. Bertinetto, J. Valmadre, S. Golodetz, O. Miksik, P. H. Torr, Staple: Complementary learners for real-time tracking, in: Computer Vision & Pattern Recognition (CVPR), 2016, pp. 1401-1409.

[64] K. Simonyan, A. Zisserman, Very Deep Convolutional Networks for Large-Scale Image Recognition, in: Computer Science, 2014.

[65] D. Kingma, J. Ba, Adam: A Method for Stochastic Optimization, in: International Conference for Learning Representations, 2015.

[66] H. Zhang, et al., "Toward Vehicle-Assisted Cloud Computing for Smartphones," IEEE Transactions on Vehicular Technology, Issue 12, Vol.64, pp. 5610-5618, Dec. 2015.

[67] H. Zhang, et al., "Interference Management for Heterogeneous Network with Spectral Efficiency Improvement," IEEE Wireless Communications Magazine, Issue 2, Vol. 22, pp. 101-107, April 2015.

[68] X. Du, M. Zhang, K. Nygard, S. Guizani, and H. H. Chen, "Self-Healing Sensor Networks with Distributed Decision Making," International Journal of Sensor Networks, Vol. 2, Nos. 5/6, pp. 289 –298, 2007.

[69] L. Wu, et al., "MobiFish: A Lightweight Anti-Phishing Scheme for Mobile Phones," in Proc. of the 23rd International Conference on Computer Communications and Networks (ICCCN), Shanghai, China, August 2014.

[70] X. Du, Y. Xiao, M. Guizani, and H. H. Chen, "An Effective Key Management Scheme for Heterogeneous Sensor Networks," Ad Hoc Networks, Elsevier, Vol. 5, Issue 1, pp 24–34, Jan. 2007.

[71] Y. Xiao, et al., "Internet Protocol Television (IPTV): the Killer Application for the Next Generation Internet," IEEE Communications Magazine, Vol. 45, No. 11, pp. 126–134, Nov. 2007.